\documentclass{article}
\usepackage{ijcai25}

\newcommand{\Pmat}[0]{\ensuremath{{\bf P}} }

\newcommand{\Tmat}[0]{\ensuremath{{\bf T}} }

\newcommand{\xv}[0]{\ensuremath{\boldsymbol{x}} }
\newcommand{\yv}[0]{\ensuremath{\boldsymbol{y}} }
\newcommand{\zv}[0]{\ensuremath{\boldsymbol{z}} }

\newcommand{\Iv}[0]{\ensuremath{\boldsymbol{I}} }

\newcommand{\alphav}[0]{\ensuremath{\boldsymbol{\alpha}} }
\newcommand{\betav}[0]{\ensuremath{\boldsymbol{\beta}} }

\newcommand{\thetav}[0]{\ensuremath{\boldsymbol{\theta}} }

\def\bb{\textcolor{blue}}

\usepackage{times}
\usepackage{soul}
\usepackage{url}
\usepackage[hidelinks]{hyperref}
\usepackage[utf8]{inputenc}
\usepackage[small]{caption}
\usepackage{graphicx}
\usepackage{amsmath}
\usepackage{amsthm}
\usepackage{booktabs}
\usepackage{algorithm}
\usepackage{algorithmic}
\usepackage[switch]{lineno}

\usepackage[utf8]{inputenc} % allow utf-8 input
\usepackage[T1]{fontenc}    % use 8-bit T1 fonts
\usepackage{url}            % simple URL typesetting
\usepackage{booktabs}       % professional-quality tables
\usepackage{amsfonts}       % blackboard math symbols
\usepackage{nicefrac}       % compact symbols for 1/2, etc.
\usepackage{microtype}      % microtypography
\usepackage{xcolor}         % colors

\usepackage{amsmath}
\usepackage{amssymb}
\usepackage{mathtools}
\usepackage{amsthm}

\usepackage[utf8]{inputenc}   % allow utf-8 input
\usepackage[T1]{fontenc}   % use 8-bit T1 fonts
\usepackage{hyperref}   % hyperlinks
\usepackage{url}   % simple URL typesetting
\usepackage{booktabs}   % professional-quality tables
\usepackage{amsfonts}   % blackboard math symbols
\usepackage{nicefrac}    % compact symbols for 1/2, etc.
\usepackage{microtype}   % microtypography
\usepackage{xcolor}   % colors

\usepackage{hyperref}
\usepackage{url}
\usepackage[utf8]{inputenc} % allow utf-8 input
\usepackage[T1]{fontenc}    % use 8-bit T1 fonts
\usepackage{hyperref}       % hyperlinks
\usepackage{url}            % simple URL typesetting
\usepackage{booktabs}       % professional-quality tables
\usepackage{amsfonts}       % blackboard math symbols
\usepackage{nicefrac}       % compact symbols for 1/2, etc.
\usepackage{microtype}      % microtypography
\usepackage{xcolor}         % colors
\usepackage{tikz} 
\usepackage{pgfplots}
\usepackage{subcaption}

\usepackage{wrapfig} % allow utf-8 input
\usepackage{graphicx}

\usepackage{amsmath}
\usepackage{amssymb}
\usepackage{algorithm} %%算法
\usepackage{algorithmic} %%算法
\usepackage{enumerate} %%用于item
\usepackage{multirow} %%表格合并行单元格
\usepackage{amsmath}

\usepackage{colortbl} 
\usepackage{xcolor}
\usepackage{array}
\usepackage{bbding}
\usepackage{enumerate}
\usepackage{bm}

\usepackage{helvet} %Required
\usepackage{courier} %Required
\usepackage{color}
\usepackage[capitalize,noabbrev]{cleveref}
\usepackage[textsize=tiny]{todonotes}
\usepackage{dsfont}

\title{\texttt{TsCA}: On the Semantic Consistency Alignment via Conditional Transport for Compositional Zero-Shot Learning}

\author{
    Miaoge Li\textsuperscript{1}, Jingcai Guo\textsuperscript{1}\thanks{Jingcai Guo is the corresponding author.}, Richard Yi Da Xu\textsuperscript{2}, Dongsheng Wang\textsuperscript{3}, \\
    Xiaofeng Cao\textsuperscript{4}, Zhijie Rao\textsuperscript{1}, Song Guo\textsuperscript{5}
    \affiliations
    \textsuperscript{1}The Hong Kong Polytechnic University~\textsuperscript{2}Hong Kong Baptist University~\\
    \textsuperscript{3}Shenzhen University~\textsuperscript{4}Jilin University~\textsuperscript{5}Hong Kong University of Science and Technology
    \emails
    Correspondence to: jc-jingcai.guo@polyu.edu.hk
}
\begin{document}

\maketitle

\begin{abstract}
Compositional Zero-Shot Learning (CZSL) aims to recognize novel \textit{state-object} compositions by leveraging the shared knowledge of their primitive components. Despite considerable progress, effectively calibrating the bias between semantically similar multimodal representations, as well as generalizing pre-trained knowledge to novel compositional contexts, remains an enduring challenge.
%significant progress, effectively calibrating the bias between semantically similar multi-modal representations and generalizing pre-trained knowledge to new compositions remains a persistent challenge.
% Despite significant progress, effectively modeling the interactions between these primitives and generalizing this knowledge to new compositions remains a persistent challenge. 
In this paper, our interest is to revisit the conditional transport (CT) theory and its homology to the visual-semantics interaction in CZSL and further, propose a novel \underline{\textbf{T}}ri\underline{\textbf{s}}ets \underline{\textbf{C}}onsistency \underline{\textbf{A}}lignment framework (dubbed \texttt{TsCA}) that well-addresses these issues. 
Concretely, we utilize three distinct yet semantically homologous sets, i.e., \textit{patches}, \textit{primitives}, and \textit{compositions}, to construct pairwise CT costs to minimize their semantic discrepancies. 
To further ensure the consistency transfer within these sets, we implement a cycle-consistency constraint that refines the learning by guaranteeing the feature consistency of the self-mapping during transport flow, regardless of modality. 
Moreover, we extend the CT plans to an open-world setting, which enables the model to effectively filter out unfeasible pairs, thereby speeding up the inference as well as increasing the accuracy. 
%significantly speeding up the inference without compromising accuracy. 
%
Extensive experiments are conducted to verify the effectiveness of the proposed method\footnote{Code is available at: https://anonymous.4open.science/r/TsCA-0E7A.}.
\end{abstract}

\section{Introduction}
Identifying new concepts from a set of seen primitives is one of the fundamental challenges for AI systems to imitate the human learning process. Imagine a new cuisine—\textit{Spicy Chocolate Cake}. Although it may sound like an odd combination, one can still perceive its appearance and taste based on existing knowledge of `\textit{spiciness}' and `\textit{Chocolate}'. 
Likewise, compositional zero-shot learning (CZSL)
% \cite{misra2017red,naeem2021learning,li2020symmetry}
~\cite{misra2017red,naeem2021learning}
emerges as a compelling paradigm that seeks to endow machines with similar cognitive abilities, allowing them to understand the composed primitives (\textit{i.e.}, states and objects) during training and generalize to unseen compositions for inference. 

\begin{figure}[!t]
\centering
\includegraphics[width=0.48\textwidth]{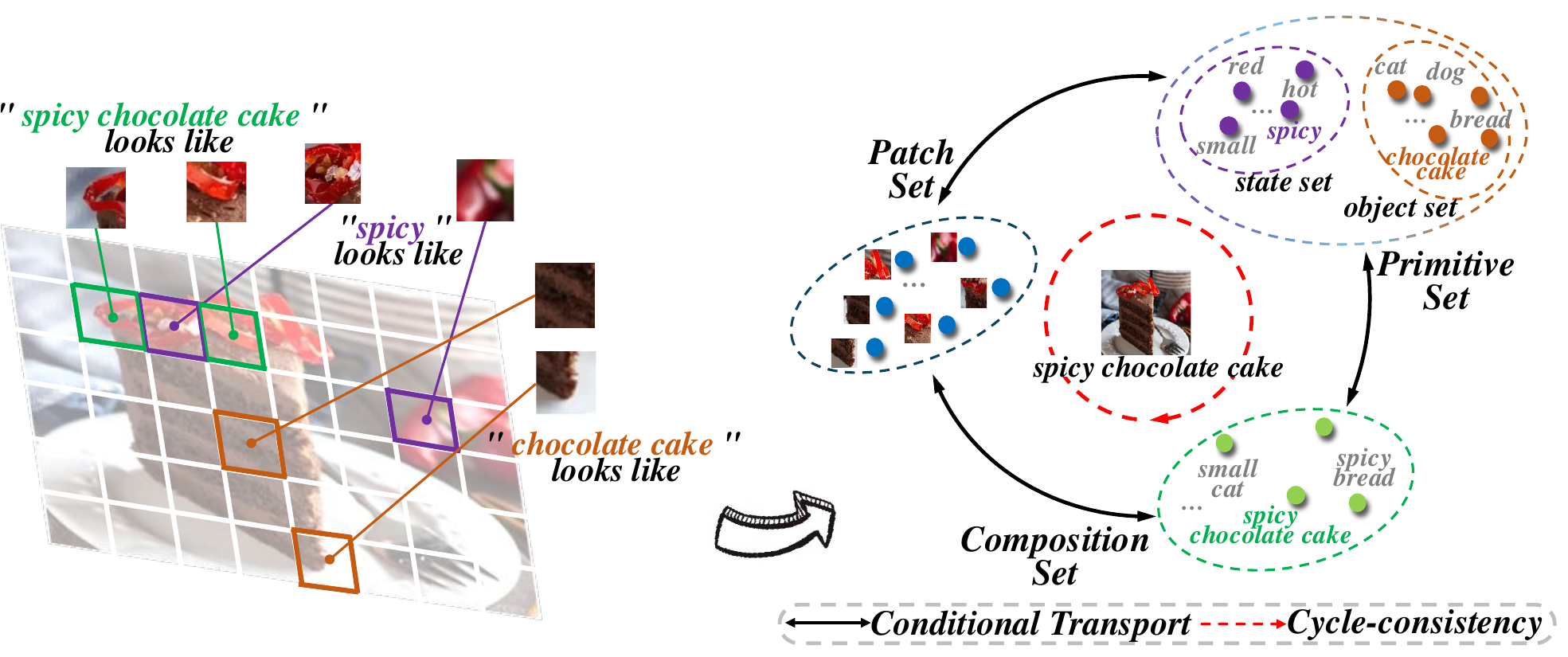}
\caption{\small{We represent each image as a set of patch embeddings and two sets of textual embeddings and employ semantic consistency conditional transport to align such cross-modal distribution trio.}}
\label{motivation}
\vspace{-7mm}
\end{figure}

Technically, studies on CZSL have revolved around achieving fine-grained alignment across the \textit{image} and composed \textit{state-object text} domains~\cite{huynh2020compositional}. Pioneer methods typically load pre-trained image encoders and word embeddings to extract multi-modal features. Various alignment tricks are then applied to constrain the shared latent space, including concept learning~\cite{xu2021zero}, geometric properties~\cite{li2020symmetry}, semantic transformation~\cite{nagarajan2018attributes}, and graph embedding~\cite{mancini2022learning}. 
Moving beyond these post-processing alignments, a series of prompt-tuning methods have been developed to explore pre-trained vision-language models (VLMs) for the CZSL scenario~\cite{xu2022prompting,nayak2022learning}, achieving state-of-the-art prediction performance. Pre-trained on massive amounts of image-text pairs, VLMs (\textit{e.g.}, CLIP~\cite{radford2021learning}) show impressive abilities to match the visual image with its textual label (prompt) in the multimodal space. Those models often employ a multi-path paradigm for fine-grained textual features, \textit{e.g.}, one composition and two primitive branches. One of the core challenges ensues: \textit{How to align the image with such multiple textual labels}? Recent studies attempt to address it from different fields, such as hierarchical prompt searching~\cite{wang2023hierarchical}, decomposed feature fusion~\cite{lu2023decomposed}, and multi-step observation~\cite{li2023compositional}. Although showing attractive results, these methods focus primarily on \textit{image-to-composition} (or \textit{to-primitive}) alignments, ignoring intrinsic relationships within the composition and its primitives. This may fail to capture semantic consistency among the \textit{image-composition-primitive} interactions, leading to suboptimal alignments.

To address the above issues, this paper proposes the \underline{\textbf{T}}ri\underline{\textbf{s}}ets \underline{\textbf{C}}onsistency \underline{\textbf{A}}lignment framework (dubbed \texttt{TsCA}), which is built upon a novel Consistency-aware Conditional Transport (CCT) 
% theory
derived from the new view of CZSL. As shown in Fig.~\ref{motivation}, we represent an image in three directions, \textit{i.e.}, $\Pmat_1$, a distribution over all visual patches; $\Pmat_2$, a distribution over the composition set; and $\Pmat_3$, a distribution over the primitive set. Concretely, $\Pmat_1$ captures the detailed visual features of an image, while $\Pmat_2$ and $\Pmat_3$ denote the global and local textual concepts of the same content. Therefore, the task of CZSL can be viewed as aligning these three discrete distributions as closely as possible. Accordingly, it is indeed key to properly measure the distance between empirical distributions with different supports. 
Fortunately, conditional transport (CT), well-examined in recent research, offers a bidirectional measurement of the distance from one distribution to another, given the cost matrix. 
Inheritedly, \texttt{TsCA} extends the minimization of CT to the alignment of our cross-modal distribution trio.
%It is natural for us to develop a new alignment framework based on the minimization of CT.

Specifically, \texttt{TsCA} gracefully facilitates the matching of \textit{image-composition-primitive} triplets by meticulously crafting three pairs of CTs, thereby aligning the rich tapestry of cross-modal and intra-modal semantics. First, $\mathcal{CT}(\Pmat_1, \Pmat_2)$ measures the transport distance between the visual patch set and the composition set. On the one hand, it helps the label be transported to the compositional visuals with higher probabilities, and on the other hand, it regularizes the finetuning for better cross-modal alignments. Next, $\mathcal{CT}(\Pmat_1, \Pmat_3)$ focuses on discovering the corresponding attribute and object patches from the disentangle perspective. Last, unlike the above two, $\mathcal{CT}(\Pmat_2, \Pmat_3)$ aims at intra-modal interactions, which are often overlooked by previous studies. By minimizing the transport distance between the composition and its primitives explicitly, the textual outputs are assumed to present high semantic coherence. \textit{E.g.}, the composition vectors are closer to their primitives in the embedding space. Moreover, the learned transport plan in $\mathcal{CT}(\Pmat_2, \Pmat_3)$ further provides an efficient tool to filter out unfeasible compositions, which benefits the open-world setting.

More importantly, as discussed above, \texttt{TsCA} formulates the CZSL as the alignment of three distributions and seeks to minimize their transport costs. 
%of image-composition-primitive triplets. 
An intuitive solution is to run pairwise transportation and optimize the above three CT costs. Unfortunately, this case could not model the complex relations among these sets. To capture deeper interactions, we extend the naive CT to a well-designed consistency-aware CT (CCT) for the CZSL triplet case. 
%develop consistency-aware CT (CCT), an extension of CT to the triplet case. 
Motivated by the fact that these three distributions describe the same semantics, CCT regularizes the product of three transport plans as a diagonal matrix from clockwise direction (shown in Fig.~\ref{motivation}). In other words, the composition label will return to itself after a cycle of transport, which ensures semantic consistency across sets. 
%
% Meanwhile, by fully utilizing the symmetry of CT, \textit{ i.e.}, $\mathcal{CT}(\Pmat_i, \Pmat_j) = \mathcal{CT}(\Pmat_j, \Pmat_i)$, \texttt{TsCA} provides a holistic estimation of the alignment distance of the triplets. 
%and we empirically find this helps better alignments. 

In summary, our contributions are three-fold:

\begin{itemize}
    \item We formulate the CZSL task as a transport problem and propose \underline{\textbf{T}}ri\underline{\textbf{s}}ets \underline{\textbf{C}}onsistency \underline{\textbf{A}}lignment model (\texttt{TsCA}), which views an image as three discrete distributions over the patch, composition, and primitive spaces and tries to explore fine-grained alignments between these triplets.
    \item We develop consistency-aware CT (CCT), considering the semantic consistency across the image-composition-primitive sets, improving the alignment robustness.
    \item Extensive comparisons and ablations on three benchmarks demonstrate the effectiveness of the proposed \texttt{TsCA} with competitive performance on all settings.

\end{itemize}

\section{Related Work}

\subsection{Compositional Zero-Shot Learning} CZSL\cite{mancini2021open} is a specialized case of zero-shot learning (ZSL)~\cite{liu2023simple}. Given the same set of objects with associated states, it focuses on recognizing unseen state-object compositions by learning from seen compositions. 
In general, traditional CZSL can be divided into two streams, \textit{i.e.,} compositional classification and simple primitive classification. The former directly predicts compositions by aligning visual features and composed labels in a shared space, or resorting to a graph network for contextuality modeling~\cite{tanwisuth2023pouf,saini2022disentangling,li2022siamese}. Conversely, the latter
identifies states and objects independently and constructs the joint compositional probability distribution. They either disregard the contextuality between primitives or impose training-specific correlations that are detrimental to generalization. 

% Recently, large-scale vision-language pre-trained
%  models(VLMs) such as CLIP have showcased outstanding performance on a wide spectrum of tasks in open-world visual concept learning. While these models acquire generalized representations, effectively adapting them to CZSL remains challenging due to the lack of sufficient supervision concerning states and their combinations with different objects. Fortunately, by prompt learning, the domain shift between the pretrained task and the downstream task is reduced, making it easier to adapt the pretrained knowledge to CZSL.
Recently, equipped with prompt learning, large-scale VLMs like CLIP are empowered to adapt to CZSL by reducing domain shift and leveraging pre-trained knowledge
% \cite{anonymous2024learning}
.
% CSP \cite{nayak2022learning} first introduces compositional soft prompting which treats the  primitive concepts as part of
% the prompt to be soft like \textit{a photo of [state] [object]}. 
% \cite{xu2022prompting}
%  additionally incorporates the prefix component into the learnable soft prompt to enhance generalization ability, particularly for a broader dataset. 
 % Further, \cite{lu2023decomposed}  initially incorporates a cross-modal decomposed fusion module between the language and image domains, followed by fusing the decomposed features to enhance the recognition of unseen compositions in the pair space. In this study, we convert the CZSL task into a hierarchical CT problem and align branch-specific prompt representations with visual features by minimizing the bidirectional transport cost.
Further, researchers attempt to explore merging those two typical paradigms to create an integrated multi-path paradigm. For instance, HPL~\cite{wang2023hierarchical} 
learns three hierarchical prompts by explicitly fixing
the unrelated word tokens in the three embedding
spaces at different levels. 
Troika~\cite{huang2024troika} effectively aligns the branch-specific prompt representations and decomposed visual features with a cross-modal traction module.
% \cite{li2024context} introduces the context-based and diversity-driven specificity learner that prioritizes specific attributes for CZSL. 
Our \texttt{TsCA} aligns closely with this paradigm, although with a greater emphasis on direct inquiries into semantic alignment within distributions. 

\subsection{Conditional Transport}
% Currently, optimal transport(OT) acts as an efficient tool to measure the distance between two discrete distributions\cite{villani2009optimal,redko2019optimal,zhao2018label,chen2020graph,lee2019hierarchical}. 
Recently, CT has acted as an efficient tool to measure the distance between two discrete distributions~\cite{zheng2021exploiting}. 
Through learning 
%an adaptive transport plan to align features, OT achieves fine-grained matching across two domains. However, it inevitably suffers from a high computational cost since it needs to optimize the transport plan via Sinkhorn algorithm\cite{cuturi2013sinkhorn} in a iterative manner. To solve this issue, Conditional Transport (CT) 
% considers 
bidirectional transport plans based on semantic similarity between samples from both source and target distributions, CT achieves fine-grained matching with the mini-batch optimization. More importantly, CT can integrate seamlessly with deep-learning frameworks and its effectiveness has been widely demonstrated in various applications~\cite{li2023patchct,wang2022representing,tanwisuth2023pouf}. For example, \cite{tian2023prototypes} propose a novel CT-based imbalanced transductive few-shot learning model to fully exploit unbiased statistics of imbalanced query samples. Similarly, \cite{tanwisuth2021prototype} employs a probabilistic bidirectional transport between target features and class prototypes for unsupervised domain adaptation, showcasing robustness against class imbalance and facilitating domain adaptation without direct access to the source data. 
For the first time, through empirical demonstration, we establish that CT, boasting attributes like lower complexity and better scalability, is equally viable for CZSL.

\begin{figure*}[!t]
\centering
\includegraphics[width=0.95\textwidth]{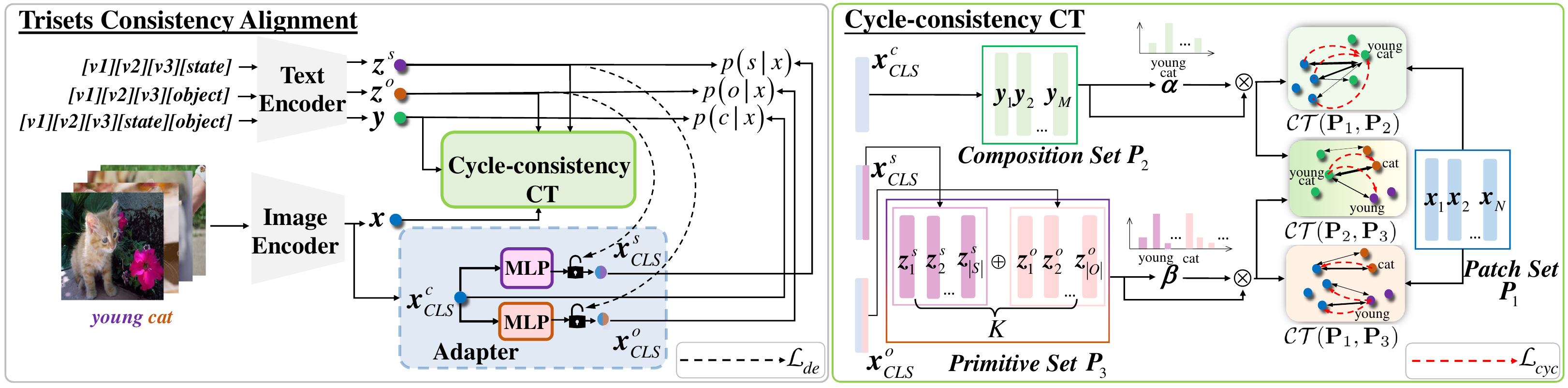}
\caption{\small{The overall framework of the proposed \texttt{TsCA}~(zoom-in for more details). }}
\label{framework}
\vspace{-2mm}
\end{figure*}

\section{Methodology}
The proposed model aims to solve the CZSL from fine-grained alignments under the CT framework, where images are viewed as three discrete distributions over the image, composition, and primitive spaces. A novel consistency-aware CT is further developed to explore deeper interactions of these three domains. In this section, we start with the task of CZSL, and then introduce how to formulate CZSL as a CT problem in detail. The framework of our approach is shown in Fig.~\ref{framework}.

\vspace{-2mm}
\subsection{Problem Formulation}
Given state set $\mathcal{S}=\{ s_0, s_1,..., s_{|\mathcal{S}|} \} $ together with object set $\mathcal{O}=\{ o_0, o_1,..., o_{|\mathcal{O}|} \} $ 
% as the primitive concepts of CZSL
, we can define the label space with their Cartesian product, $\mathcal{C}=\mathcal{S} \times \mathcal{O}$. Then $\mathcal{C} $ can be divided into two disjoint label subsets such that $\mathcal{C}^{se} \in \mathcal{C}$, $\mathcal{C}^{us} \in \mathcal{C}$, and  $\mathcal{C}^{se} \cap \mathcal{C}^{us} = \emptyset$ where $\mathcal{C}^{se}$ and $\mathcal{C}^{us}$ are the set of the seen and unseen classes respectively. Specifically, during the training phase, $\mathcal{C}^{se}$ are used to train a discriminative model $\mathcal{M}:\mathcal{X}\overrightarrow \mathcal{C}^{se}$ from the input image space to candidate composition label set. At inference time, the model is expected to predict unseen compositions in the test sample space, i.e.,$\mathcal{M}:\mathcal{X}\overrightarrow \mathcal{C}^{test}$. In this paper, we follow the setting of Generalized ZSL \cite{Xian2017ZeroShotLC}, considering testing samples contain both seen and unseen compositions. In general, in the closed-world evaluation, only the known composition space of test samples is required as $\mathcal{C}^{test} = \mathcal{C}^{se} \cup \mathcal{C}^{us}$. For the open-world evaluation \cite{Karthik2022KGSPKG}, the model has to consider all possible permutations of the state-object pairs, \textit{i.e.}, $\mathcal{C}^{test} = \mathcal{C}$.
% and $\mathcal{C}^{u} = \mathcal{C} \cup \mathcal{C}^{s}$. 

\vspace{-2mm}
\subsection{Image as Three Sets}
Built upon the pre-trained CLIP and with the soft prompt tuning technique, our \texttt{TsCA} represents an image as three sets: visual patch set 
% $\Pmat_1$
, textual composition set 
% $\Pmat_2$
, and primitive set 
% $\Pmat_3$
. These sets capture the multimodal features of the same content from different semantic domains, acting as a fundamental role in the fine-grained alignments.

% \vspace{-2mm}
\paragraph{Patch Set.} For an input image $x$, CLIP first splits it into $N$ non-overlapping patches evenly and then feeds them (with a [CLS] token inserted) into the image encoder to extract the embedding of the [CLS] token $\xv^c_{CLS}$ and the patch features $\xv=\{\xv_n | _{n=1}^{N}  \}\in \mathbb{R}^{d \times N }$, where $\xv_n \in \mathbb{R}^{d}$ denotes the embedding of patch $n$ with the embedding dimension being $d$. Naturally, the discrete probability distribution over the patch set can be formulated as:
\begin{equation} \label{comp_set}
    \Pmat_1 = \sum_{n=1}^N \theta_n \delta_{\xv_n}, \quad \theta_n=\frac{1}{N},
\end{equation}
where $\delta$ denotes the Dirac delta function. We view all the patches equally and employ the Uniform distribution to model the patch weight $\thetav$. Note that $\Pmat_1$ collects the detailed visual features of the local region, thereby bringing benefits to discriminative representation learning, especially when different concepts require emphasis on distinct areas within the input images. Besides, the [CLS] visual token $\xv_{CLS}^c$ is also learned, serving as the global representation, which is often used as the image feature to downstream tasks \cite{zhou2022learning}. 
Here, we consider it as the visual prior that constructs subsequent textual sets. 
% Two distinct adapters are employed to further transform $\xv^{c}_{CLS}$ to state representation $\xv^{s}_{CLS}$ and object representation $\xv^{o}_{CLS}$, allowing the derivation of unique visual representations for states and objects respectively. Once learned, such disentangled global visual representations will be treated as primitive navigator prior information to guide subsequent CT learning.

\paragraph{Composition Set.}
In the textual domain, we build composition labels into a learnable soft prompt \textit{\underline{[v1][v2][v3][state][object]}}, where \textit{[v]} is the prefix vectors, and \textit{[state][object]} denote the state and object name, respectively. Then, the prompt will be fed into the text encoder to obtain textual representations $\yv = \{ \yv_m | _{m=1}^M \}\in \mathbb{R}^{d \times M}$, where $M$ is the number of training compositions.
To embrace the variety of content and minimize the cross-modal disparities, we follow previous work~\cite{huang2024troika} and incorporate a residual component obtained through a cross-attention mechanism~\cite{vaswani2017attention} with patch embeddings (we here still use $\yv$ as the output compositions):
\begin{equation} \label{cross-atten}
    \yv = \text{Cross-Att}(\yv^{in}, \xv, \xv) + \yv^{in},
\end{equation}
where $\yv^{in}$ denotes the input compositions. \text{Cross-Att} is the cross-attention layer
with the query $\yv^{in}$, key $\xv$, and value $\xv$ as inputs, and outputs the fused features. As a result, the composition set can be viewed as:
\begin{equation} \label{comp_set}
    \Pmat_2 = \sum_{m=1}^M \alpha_m \delta_{\yv_m}, \quad \alphav = \sigma(\yv^T\xv^c_{CLS}),
\end{equation}
where $\sigma$ denotes the softmax function. We calculate the composition weights $\alphav$ via the semantic similarity of the composition label and visual feature. This helps $\Pmat_2$ to focus on compositions that describe the input image well. 

\paragraph{Primitive Set.}
Unlike the composition set that focuses on global textual features, the primitive set aims to explore the disentangled state and object representations. Motivated by the recent multi-path paradigms, we learn the corresponding primitive representations through \textit{\underline{[v1][v2][v3][state]}} and \textit{\underline{[v1][v2][v3][object]}}, with the outputs denoted as $\zv^s \in \mathbb{R}^{d \times {|\mathcal{S}|}}$ and $\zv^o \in \mathbb{R}^{d \times {|\mathcal{O}|}}$ respectively. The corresponding probability weights are calculated as:
\begin{equation}
    \betav^s = \sigma(({\zv^s})^T \xv^s_{CLS}), \quad \betav^o = \sigma(({\zv^o})^T \xv^o_{CLS}),
\end{equation}
where $\xv^s_{CLS}$ and $\xv^o_{CLS}$ are two adapted visual features, which can be obtained via a lightweight adapter $g$:
\begin{equation}
    [\xv^s_{CLS}, \xv^o_{CLS}] = g(\xv^c_{CLS}),
\end{equation}
where $g$ takes the image feature $\xv^c_{CLS}$ as input, and outputs the state/object-relevant features, deriving unique visual representations. To complete the primitive set, we concatenate the feature points from both state and object labels as $\zv = \{ \zv_{k} | _{k=1}^{K} \}\in \mathbb{R}^{d \times K}$, where $K=|\mathcal{S}| + |\mathcal{O}|$ is the total number of state and object labels. Like compositions that fuse the patch features with Eq.~\ref{cross-atten}, $\zv$ is fed into the same cross-attention layer:
\begin{equation} 
    \zv = \text{Cross-Att}(\zv^{in}, \xv, \xv) + \zv^{in}.
\end{equation}
Finally, the primitive set can be expressed as:
\begin{equation}\label{comp_set}
    \Pmat_3 = \sum_{k=1}^{K} \beta_k \delta_{\zv_k}, \quad 
    \betav = \sigma (\betav^s \oplus \betav^o),
\end{equation}
where $\oplus$ denotes the concatenation operation.
% The probability weight of the primitive points is obtained by normalizing the concatenated similarity.
Together with $\Pmat_2$, these two sets provide primitive-composition textual knowledge for downstream alignment tasks.  

% \vspace{-2mm}
\subsection{Semantic Consistency Alignment}
Based on the three carefully constructed sets of the input image, our \texttt{TsCA} formulates the CZSL task as fine-grained alignments under the consistency-aware CT framework, which consists of pairwise CT distance and cycle consistency regularization. 
\subsubsection{Pairwise CT Distance.}
Given the source distribution $\Pmat_1$ and the target distribution $\Pmat_2$, CT measures the distance by calculating the total transport costs bidirectionally, leading to the forward CT and backward CT, respectively. Denoting $c(\xv_n, \yv_m) \geq 0$ as a cost function to define the difference between points $\xv_n$ and $\yv_m$, the forward CT is measured as the expected transport cost of transporting all points from $\Pmat_1$ to the 
target $\Pmat_2$ 
% set
, and the backward CT reverses the transport direction. Mathematically, the CT distance between $\Pmat_1$ and $\Pmat2$ can be expressed as:
% \begin{equation}

% \begin{align} \label{ct}
%     &\mathcal{CT}_{\Cmat_{12}}(\Pmat_1, \Pmat_2)= \min _{ \overrightarrow{\Tmat} \overleftarrow{\Tmat}}  (\sum_{n,m} \overrightarrow{t}_{nm} c_{nm} + \sum_{m,n} \overleftarrow{t}_{mn} c_{mn}),\\
%      &\mathrm{s.t.} 
% \overrightarrow{\Tmat} \mathds{1}^{N}=\thetav, \overleftarrow{\Tmat}  \mathds{1}^M=\alphav, \notag
% \end{align}

\begin{align} \label{ct}
    &\mathcal{CT}(\Pmat_1, \Pmat_2)= \min _{ \overrightarrow{\Tmat} \overleftarrow{\Tmat}}  (\sum_{n,m} \overrightarrow{t}_{nm} c_{nm} + \sum_{m,n} \overleftarrow{t}_{mn} c_{mn}),\\
     &\mathrm{s.t.} 
\overrightarrow{\Tmat} \mathds{1}^{N}=\thetav, \overleftarrow{\Tmat}  \mathds{1}^M=\alphav, \notag
\end{align}
where we employ the cosine distance as the cost function, \textit{i.e.}, the closer the patch $\xv_n$ and composition $\yv_m$ are, the lower the transport cost. $\mathds{1}^{M}$ is the $M$ dimensional vector of ones.
$\Tmat$ is called the transport plan, which is to be learned to minimize the total distance. The forward transport plan, \textit{e.g.}, $\overrightarrow{t}_{nm}$ in $\overrightarrow{\Tmat}$ describes the transport probability from the source point $\xv_n$ to target point $\yv_m$: 
\begin{equation}\label{transport plan}
\overrightarrow{t}_{nm} = \theta_n \frac{\alpha_m e^{s_{\psi}(\xv_n,\yv_m)}}{\sum_{m^{\prime}=1}^{M} \alpha_{m^{\prime}} 
e^{s_{\psi}(\xv_n,\yv_{m^{\prime}})}},
\end{equation} 
where $s_{\psi}$ denotes the similarity function with learnable parameter $\psi$, and we specify it as $s_{\psi}(\xv_n,\yv_m) = \frac{\xv_n^T \yv_m}{\text{exp}(\psi)}$. That is, the patch $\xv_n$ will transport with higher probability if the composition $\yv_m$ shares similar semantics with it. 
Similarly, we have the backward transport expressed as:
\begin{equation}\label{transport plan'}
\overleftarrow{t}_{mn} = \alpha_m \frac{\theta_n e^{s_{\psi}(\yv_m,\xv_n)}}{\sum_{n^{\prime}=1}^{N} \theta_{n^{\prime}} e^{s_{\psi}(\yv_m,\xv_{n^{\prime}})}}.
\end{equation}
More interestingly, the definition in Eq.~\ref{transport plan}-~\ref{transport plan'} naturally satisfies the constraint in Eq.~\ref{ct}, simplifying the CT optimization.

By combining $\Pmat_1$, $\Pmat_2$, and $\Pmat_3$ pairwise, we extend CT to the CZSL scenario and denote the total CT distances as:
% \begin{equation} \label{ct}
%   \text{CT}_{\Cmat} = 
%     \text{CT}_{\Cmat_{12}}(\Pmat_1, \Pmat_2)+    \text{CT}_{\Cmat_{23}}(\Pmat_2, \Pmat_3) +  \text{CT}_{\Cmat_{31}}(\Pmat_3, \Pmat_1) 
% \end{equation}

% \begin{equation} \label{ct_distance}
%   \mathcal{CT} = 
%     \mathcal{CT}_{\Cmat_{12}} (\Pmat_1, \Pmat_2)+    \mathcal{CT}_{\Cmat_{13}} (\Pmat_1, \Pmat_3) +  \mathcal{CT}_{\Cmat_{23}}(\Pmat_2, \Pmat_3),
% \end{equation}
\vspace{-4pt}
\begin{equation} \label{ct_distance}
  \mathcal{CT} =  \mathcal{CT} (\Pmat_1, \Pmat_2)+    \mathcal{CT} (\Pmat_1, \Pmat_3) +  \mathcal{CT}(\Pmat_2, \Pmat_3),
\end{equation}
\vspace{3pt}
where the first two terms facilitate cross-domain alignments but focus on different semantic levels.
% $\mathcal{CT}_{\Cmat_{12}}$ 
$\mathcal{CT} (\Pmat_1, \Pmat_2)$
takes the patch set and composition set as inputs and aims to find compositional visuals that match its textual label. While 
% $\mathcal{CT}_{\Cmat_{13}}$ 
$\mathcal{CT} (\Pmat_1, \Pmat_3)$
pays more attention to primitive alignment. Intuitively, the primitive set provides decoupled textual guidance, and it helps the model extract patches that contain the state or object-relevant visuals, which will improve the fine-grained predictions. Moving beyond the vision-language alignments, the last term attempts to explore the intra-modal interactions. 
% $\mathcal{CT}_{\Cmat_{23}}$ 
$\mathcal{CT} (\Pmat_2, \Pmat_3)$ optimizes the composition and its primitives to be semantically close in the embedding space, showing linguist coherence.
\subsubsection{Cycle Consistency Regularization.}
Due to the independent operation in Eq.~\ref{ct_distance}, pairwise CTs may lead to potential inconsistencies and inefficiencies, as they fail to fully capture the global relationships among all sets involved. Cycle consistency, which has been extensively explored in domains such as image matching \cite{bernard2019hippi}, multi-graph alignment \cite{wang2021neural}, and 3D pose estimation \cite{dong2019fast}, offers a natural solution to this problem. By enforcing a closed-loop structure, cycle consistency can enhance the coherence of semantic relationships across all sets, ensuring more consistent and efficient alignment.
Here, we denote the probability of the composition set being self-mapping back to its initial state as:
\begin{equation} \label{comp_set}
    {\Tmat}_{22} = \overleftarrow{\Tmat}_{21} \cdot \overrightarrow{\Tmat}_{13} \cdot \overleftarrow{\Tmat}_{32},
\end{equation}
where $\overleftarrow{\Tmat}_{21}$ from $\mathcal{CT}(\Pmat_1, \Pmat_2)$ denotes the transport plan from $\Pmat_2$ to $\Pmat_1$.
This guarantees each label is transferred with a probability of 1 during transportation among three sets. 
For a given input, the corresponding point from the composition set first interacts with key patches in the patch set, then moves to the relevant state and object points within the primitive set, and finally returns to the composition set.
Accordingly, we derive the cycle-consistency constraint by:
\begin{equation} \label{cyc}
    \mathcal{L}_{cyc}= \sum_{m=1}^M 
    \yv^c_m
    (\Tmat_{22} - \Iv),
\end{equation}
where $\yv^c_m$ denotes the one-hot binary label vector of the input image.

For one thing, such a closed-loop transport constraint establishes a close link between independent pairwise CTs, which supports maintaining semantic consistency throughout the CT transport process. For another, it facilitates more robust composition representation learning, enhancing the model's accuracy. 

% \rr{How to merge multiple losses reasonably?
% disentangled loss/cross-entropy loss/ct loss/consistency loss}
\subsubsection{Primitive Decoupler.}
In light of the intricate entanglement between attributes and objects within an image, we devise a decoupling loss with the idea that visual representation $\xv^{s}_{CLS}$ and $\xv^{o}_{CLS}$ can be seen as state-expert and object-expert if their paired sub-concepts 
% in the composition 
fail to be inferred:
\begin{equation} \label{decoupler}
%   \mathcal{L}_{de} = 
% |cos(\xv^{CLS}_s,\fv^o_{gt})|+|cos(\xv^{CLS}_o,\fv^s_{gt})|
  \mathcal{L}_{de} = 
\left\| cos(\xv^s_{CLS},\zv^o_{gt}) \right\|+
\left\| cos(\xv^o_{CLS},\zv^s_{gt}) \right\|,
\end{equation}
where $\zv^o_{gt}$ and $\zv^s_{gt}$
are textual representations for the ground-truth object class and attribute class, respectively. By mitigating the entanglement among visual representations, \texttt{TsCA} enhances its capacity to pinpoint correlative image representations that align with specific knowledge, thereby offering guidance on the textual sets prior and playing a complementary role in deriving a more accurate composition during inference.

\subsection{Tranining and Inference}
\subsubsection{Training Objectives.}
Recalling that the probability weights $\alphav$, $\betav^s$, and $\betav^o$, calculated during the construction of the three sets, capture the semantic similarity between the visual image and the corresponding textual features, this naturally defines the probability for predicting the labels of state $s$, object $o$, and composition $c$ for the image $x$:
\begin{equation}
    p(s|x) = \betav^s, \quad p(o|x) = \betav^o, \quad p(c|x) = \alphav
\end{equation}
Then the classification losses are given by:
\begin{equation}
\left\{\begin{matrix}
  \mathcal{L}_s = -  \frac{1}{|\mathcal{X}|}{\sum_{\hat{x}\in \mathcal{X}}\text{log}p(s|x)}\\ 
  \mathcal{L}_o = -  \frac{1}{|\mathcal{X}|}{\sum_{\hat{x}\in \mathcal{X}}\text{log}p(o|x)}\\ 
  \mathcal{L}_c = -  \frac{1}{|\mathcal{X}|}{\sum_{\hat{x}\in \mathcal{X}}\text{log}p(c|x)}  
\end{matrix}\right..
\end{equation}
Let $\mathcal{L}_{base} = \mathcal{L}_{c} + 
\mathcal{L}_{s} +
\mathcal{L}_{o}$, the overall training loss is defined as follows:
\begin{equation} \label{loss}
  \mathcal{L} = \lambda_0 \mathcal{L}_{base} + 
  \lambda_1 \mathcal{CT} +
  \lambda_2
  \mathcal{L}_{cyc} +
  \lambda_3
  \mathcal{L}_{de},
\end{equation}
where $\lambda_{\text{-}}$ are hyper-parameters to balance the losses. Like the previous works, the first term is our base classification loss. It will be used to predict the final label via a combined strategy during the inference. In addition, it helps to construct increasingly coherent textual sets as the loss decreases. The last three terms act as semantic regularization, which guides the learning process from various domain experts. 
\subsubsection{Inference.}
With multi-path union, the prediction results of states
and objects can be incorporated to assist the composition
branch. Formally, the integrated composition probability can be denoted as:
\begin{equation} \label{inference}
   \tilde{p}(c|x) =  \gamma p(c|x) + (1-\gamma)(p(s|x) \times p(o|x)),
\end{equation}
where $\gamma$ balances the contributions of primitive prediction and composition prediction in multi-path learning. 
% The composition with the highest probability is predicted. 

Additionally, we apply a single CT computation to filter out infeasible compositions that might be present in the open-world setting. Concretely, we calculate the bidirectional transport plans between primitives and compositions as their feasible scores, and discarded less relevant pairs by a threshold $T$ in : 
\begin{equation} 
\label{filter}   
 % \mathcal{C}^{test^{'}}= \left\{c = (s,o):{\overrightarrow{t}_{c \rightarrow s} } {\overrightarrow{t}_{c \rightarrow o}} +
 % {\overrightarrow{t}_{s \rightarrow c}}
 % {\overrightarrow{t}_{c \rightarrow o}}
 %  {\overrightarrow{t}_{s \rightarrow c}}
 % \right\}
  \mathcal{C}^{test^{'}}= \left\{c = (s,o):{\overrightarrow{t}_{cs} } {\overrightarrow{t}_{co}} +
 {\overleftarrow{t}_{sc}}
 {\overleftarrow{t}_{oc}} < T
 \right\}.
\end{equation}
This strategy reduces the search space and increases performance simultaneously.
% This strategy reduces the search space and increases performance,retains fewer but more relevant pairs, leading to superior HM and AUC.

\section{Experiment}
\subsection{Experimental Settings}

\textbf{Datasets.} The proposed \texttt{TsCA} is evaluated on three real-world CZSL benchmark datasets: MIT-states\cite{isola2015discovering}, UT-Zappos\cite{yu2014fine} and C-GQA \cite{naeem2021learning}. 
MIT-States comprises images of naturally occurring objects, with each object characterized by an accompanying adjective description. It comprises 53,753 images depicting 115 states and 245 objects. 
UT-Zappos is a fine-grained dataset consisting of different kinds of shoes with texture attributes, totaling 16 states and 12 objects. 
C-GQA, derived from the Stanford GQA dataset~\cite{hudson2019gqa}, features 453 states and 870 objects with over 9,500 compositions, making it the most pairs dataset for CZSL. 
We follow the split suggested by the previous
work~\cite{purushwalkam2019task} to ensure fair
comparisons and detailed statistics can be found in 
% Tab.~\ref{tab:datasets}.
Appendix.

% \begin{table}[h]
% % \begin{table}[!ht]
% \centering
% \caption{Statistics of datasets}
\vspace{2mm}
% \label{tab:datasets}
% \resizebox{\columnwidth}{!}{ % This command will resize the table to fit the column width
% \begin{tabular}{|l|cc|ccc|ccc|}
% \hline
% \textbf{Dataset} & \multicolumn{2}{c|}{\textbf{Train}} & \multicolumn{3}{c|}{\textbf{Validation}} & \multicolumn{3}{c|}{\textbf{Test}} \\
%  & \textbf{|Y\textsubscript{s}|} & \textbf{|X|} & \textbf{|Y\textsubscript{s}|} & \textbf{|Y\textsubscript{u}|} & \textbf{|X|} & \textbf{|Y\textsubscript{s}|} &
%  \textbf{|Y\textsubscript{u}|} &\textbf{|X|} \\
% \hline
% MIT-States & 1262 & 30338 & 300 & 300 & 10420 & 400 &400 & 12995 \\
% UT-Zappos & 83 & 22998 & 15 & 15 & 3214 & 18 &18 & 2914 \\
% C-GQA & 5592 & 26920 & 1252 & 1040 & 7280 & 888 & 923 & 5098 \\
% \hline
% \end{tabular}
% }
% \end{table}
% % \vspace{-2mm}

\begin{table*}[t]
\fontsize{8}{8}\selectfont
\centering
\vspace{-2mm}
\begin{tabular}{lcccccccccccc}
\toprule
& \multicolumn{4}{c}{MIT-States} & \multicolumn{4}{c}{UT-Zappos} & \multicolumn{4}{c}{C-GQA} \\
\cmidrule(lr){2-5} \cmidrule(lr){6-9} \cmidrule(l){10-13}
\rowcolor{gray!20} % 设置第一行的背景颜色为灰色
Method & S & U & H & AUC & S & U & H & AUC & S & U & H & AUC\\
\midrule
% AoP & 14.3 & 17.4 & 9.9 & 1.6 & 59.8 & 54.2 & 40.8 & 25.9 & 17.0 & 5.6 \\
% LE+ & 15.0 & 20.1 & 10.7 & 2.0 & 53.0 & 61.9 & 41.0 & 25.7 & 18.1 & 5.6 \\
% ... & ... & ... & ... & ... & ... & ... & ... & ... & ... & ... \\
% CompCos & 25.3 & 24.6 & 16.4 & 4.5 & 59.8 & 62.5 & 43.1 & 28.1 & 28.1 & 11.2 \\
% CGE & 28.7 & 25.3 & 17.2 & 5.1 & 56.8 & 63.6 & 41.2 & 26.4 & 28.7 & 25.3 \\
CLIP 
\cite{radford2021learning}
& 30.2 & 46.0 & 26.1 & 11.0 & 15.8 & 49.1 & 15.6 & 5.0 & 7.5 & 25.0 & 8.6 & 1.4 \\
CoOp 
\cite{zhou2022learning}
& 34.4 & 47.6 & 29.8 & 13.5 & 52.1 & 49.3 & 34.6 & 18.8 & 20.5 & 26.8 & 17.1 & 4.4 \\
PromptCompVL 
\cite{xu2022prompting}
& 48.5 & 47.2 & 35.3 & 18.3 & 64.4 & 64.0 & 46.1 & 32.2 & —— & —— & —— & ——\\
CSP 
\cite{nayak2022learning}
& 46.6 & 49.9 & 36.3 & 19.4 & 64.2 & 66.2 & 46.6 & 33.0 & 28.8 & 26.8 & 20.5 & 6.2 \\
HPL 
\cite{wang2023hierarchical}
& 47.5 & 50.6 & 37.3 & 20.2 & 63.0 & 68.8 & 48.2 & 35.0 & 30.8 & 28.4 & 22.4 & 7.2 \\
GIPCOL 
\cite{xu2024gipcol}
& 48.5 & 49.6 & 36.6 & 19.9 & 65.0 & 68.5 & 48.8 & 36.2 & 31.9 & 28.4 & 22.5 & 7.1 \\
DFSP (i2t) 
\cite{lu2023decomposed}
& 47.4 & 52.4 & 37.2 & 20.7 & 64.2 & 66.4 & 45.1 & 32.1 & 35.6 & 29.3 & 24.3 & 8.7 \\
DFSP (BiF) 
\cite{lu2023decomposed}
& 47.1 & 52.8 & 37.7 & 20.8 & 63.3 & 69.2 & 47.1 & 33.5 & 36.5 & 32.0 & 26.2 &9.9\\
DFSP (t2i) 
\cite{lu2023decomposed}
& 46.9 & 52.0 & 37.3 & 20.6 & 66.7 & 71.7 & 47.2 & 36.0 & 38.2 & 32.0 & 27.1 & 10.5\\
PLID 
\cite{bao2023prompting}
& 49.7 & 52.4 & 39.0 & 22.1 & 67.3 & 68.8 & 52.4 & 38.7 & 38.8 & 33.0 & 27.9 & 11.0 \\
Troika 
\cite{huang2024troika}
& 49.0 & \bb{53.0} & \bb{39.3} & 22.1 & 66.8 & 73.8 & 54.6 & 41.7 & 41.0 & 35.7 & 29.4 & 12.4 \\
%PLO-VLM
% \cite{li2023compositional} 
%& 49.6 & 52.7 & 39.0 & 22.2 & 67.8 & \textbf{75.6}
% & 53.1 & 42.0 & 43.9 & \bb{38.2} & \bb{32.2} & \bb{14.5}\\
%PLO-LLM
% \cite{li2023compositional} 
%& 49.6 & \bb{53.2} & 39.0 & 21.9 & 68.3 & 73.0 & 54.8 & 41.6 & \bb{44.3} & 37.9 & 31.2 & 14.3\\
CDS-CZSL
\cite{li2024context} 
& \bb{50.3} & 52.9 & 39.2 & 22.4 & 63.9 & \bb{74.8} & 52.7 & 39.5 & 38.3 & 34.2 & 28.1 & 11.1\\

Retrieval-Augmented
\cite{jing2024retrieval}
& 50.0 & \textbf{53.3} & 39.2 & \bb{22.5} & \textbf{69.4} & 72.8 & \bb{56.5} & \bb{44.5} & \textbf{45.6} & \bb{36.0} & \bb{32.0} & \bb{14.4}\\
% ... & ... & ... & ... & ... & ... & ... & ... & ... & ... & ...  & ... & ...\\
% CT-CZSL(ours) & 48.7 & 50.3 & 37.7 & 20.5 & 65.8 & 74.5 & 55.5 & 42.1 & 40.1 & 30.3 & 23.8 & 10.4\\
% Troika-CT(ours) & 49.5 & 52.9 & 38.8 & 22.1 & —— & —— & —— & —— & —— & —— & —— & ——\\
% Troika-OT(ours) & 49.3 & 53.4 & 38.9 & 22.2 & —— & —— & —— & —— & —— & —— & —— & ——\\
% Troika-CT+CC(ours) & 50.1 & 53.1 & 39.3 & 22.5 & —— & —— & —— & —— & —— & —— & —— & ——\\
% Troika-OT+CC(ours) & 49.1 & 53.5 & 38.7 & 22.0 & —— & —— & —— & —— & —— & —— & —— & ——\\
% Troika-CT+ de-bias & 51.2 & 52.9 & 39.9 & 23.0 & 69.8 & 74.9 & 56.7 & 44.4 & 40.6 & 36.2 & 30.0 & 12.6\\
\midrule
% \texttt{TsCA} & 51.2 & 52.9 & 39.9 & 23.0 & 68.6 & 74.3 & 58.4 & 45.1 & 40.6 & 36.2 & 30.0 & 12.6\\
% \texttt{TsCA} & 51.2 & 52.9 & 39.9 & 23.0 & 68.6 & 74.3 & 58.4 & 45.1 & 42.0 & 37.0 & 30.9 & 13.3\\
% \texttt{TsCA} & \textbf{51.2} & 52.9 & \textbf{39.9} & \textbf{23.0} & 68.6 & 74.3 & \textbf{58.4} & \textbf{45.1} & 41.9 & 37.9 & 31.3 & 13.6\\
\texttt{TsCA} & \textbf{51.2} & 52.9 & \textbf{39.9} & \textbf{23.0} & \bb{68.7} & \textbf{75.8} & \textbf{58.5} & \textbf{46.1} 
% & 42.5 & 37.9 & 31.8 & 14.0\\
% & 42.6 & \textbf{38.4} & \bb{32.0} & 14.2\\
% & 43.0 & \textbf{38.2} & \textbf{32.3} & 14.3\\
& \bb{43.8} & \textbf{38.9} & \textbf{33.1} & \textbf{15.2}\\

% \textbf{}

\bottomrule
\end{tabular}
\caption{
% Close-world results on MIT-States, UT-Zappos and C-GQA. \( S \) and \( U \) are the predict accuracies evaluated on seen and unseen compositions. \( H \) is the harmonic mean of \( U \) and \( S \) and AUC is the area under the curve.
CZSL comparisons in the closed-world setting.
The best results are in \textbf{bold}. The second best results are in \bb{blue}.}
\label{tab:results_cw}
\end{table*}
\vspace{-2mm}

\begin{table*}[t]
\fontsize{8}{8}\selectfont
\centering
\vspace{-2mm}
\begin{tabular}{lcccccccccccc}
\toprule
& \multicolumn{4}{c}{MIT-States} & \multicolumn{4}{c}{UT-Zappos} & \multicolumn{4}{c}{CGQA} \\
\cmidrule(lr){2-5} \cmidrule(lr){6-9} \cmidrule(l){10-13}
\rowcolor{gray!20} % 设置第一行的背景颜色为灰色
Method & S & U & H & AUC & S & U & H & AUC & S & U & H & AUC\\
\midrule
% AoP & 14.3 & 17.4 & 9.9 & 1.6 & 59.8 & 54.2 & 40.8 & 25.9 & 17.0 & 5.6 \\
% LE+ & 15.0 & 20.1 & 10.7 & 2.0 & 53.0 & 61.9 & 41.0 & 25.7 & 18.1 & 5.6 \\
% ... & ... & ... & ... & ... & ... & ... & ... & ... & ... & ... \\
% CompCos & 25.3 & 24.6 & 16.4 & 4.5 & 59.8 & 62.5 & 43.1 & 28.1 & 28.1 & 11.2 \\
% CGE & 28.7 & 25.3 & 17.2 & 5.1 & 56.8 & 63.6 & 41.2 & 26.4 & 28.7 & 25.3 \\
CLIP 
\cite{radford2021learning}
& 30.1 & 14.3 & 12.8 & 3.0 & 15.7 & 20.6 & 11.2 & 2.2 & 7.5 & 4.6 & 4.0 & 0.3 \\
CoOp 
\cite{zhou2022learning}
& 34.6 & 9.3 & 12.3 & 2.8 & 52.1 & 31.5 & 28.9 & 13.2 & 21.0 & 4.6 & 5.5 & 0.7 \\
PromptCompVL 
\cite{xu2022prompting}
& 48.5 & 16.0 & 17.7 & 6.1 & 64.6 & 44.0 & 37.1 & 21.6 & —— & —— & —— & ——\\
CSP 
\cite{nayak2022learning}
& 46.3 & 15.7 & 17.4 & 5.7 & 64.1 & 44.1 & 38.9 & 22.7 & 28.7 & 5.2 & 6.9 & 1.2 \\
HPL 
\cite{wang2023hierarchical}
& 46.4 & 18.9 & 19.8 & 6.9 & 63.4 & 48.1 & 40.2 & 24.6 & 30.1 & 5.8 & 7.5 & 1.4 \\
GIPCOL 
\cite{xu2024gipcol}
& 48.5 & 16.0 & 17.9 & 6.3 & 65.0 & 45.0 & 40.1 & 23.5 & 31.6 & 5.5 & 7.3 & 1.3 \\
DFSP (i2t) 
\cite{lu2023decomposed}
& 47.2 & 18.2 & 19.1 & 6.7 & 64.3 & 53.8 & 41.2 & 26.4 & 35.6 & 6.5 & 9.0 & 2.0 \\
DFSP (BiF) 
\cite{lu2023decomposed}
& 47.1 & 18.1 & 19.2 & 6.7 & 63.5 & 57.2 & 42.7 & 27.6 & 36.4 & 7.6 & 10.6 & 2.4\\
DFSP (t2i) 
\cite{lu2023decomposed}
& 47.5 & 18.5 & 19.3 & 6.8 & 66.8 & 60.0 & 44.0 & 30.3 & 38.3 & 7.2 & 10.4 & 2.4\\
PLID
\cite{bao2023prompting}
& 49.1 & 18.7 & 20.0 & 7.3 & 67.6 & 55.5 & 46.6 & 30.8 & 39.1 & 7.5 & 10.6 & 2.5 \\
Troika 
\cite{huang2024troika}
& 48.8 & 18.7 & 20.1 & 7.2 & 66.4 & 61.2 & 47.8 & 33.0 & 40.8 & 7.9 & 10.9 & 2.7 \\
%PLO-VLM
% \cite{li2023compositional} 
%& 49.5 & 18.7 & 20.5 & 7.4 & 68.0 & 63.5 & 47.8 & 33.1 & 43.9 & 10.4 & 13.9 & 3.9\\
CDS-CZSL
\cite{li2024context} 
& 49.4 & \textbf{21.8} & \bb{22.1} & 8.5 & 64.7 & 61.3 & 48.2 & 32.3 & 37.6 & 8.2 & 11.6 & 2.7\\
Retrieval-Augmented 
\cite{jing2024retrieval}
& 49.9 & 20.1 & 21.8 & 8.2 & 69.4 & 59.4 & 47.9 & 33.3 & \textbf{45.5} & 11.2 & \bb{14.6} & \bb{4.4}\\
% ... & ... & ... & ... & ... & ... & ... & ... & ... & ... & ... & ... & ...\\
\midrule
% \texttt{TsCA}& 50.7 & 18.9 & 20.2 & 7.5 & 67.6 & 63.5 & 52.0 & 36.1 & - & - & - & -\\

\texttt{TsCA} (w/o filter)& 
% \textbf{50.7} & 21.0 & 21.6 & 8.4 & 67.6 & 63.5 & 52.0 & 36.1
% \textbf{50.7} & 21.0 & 21.6 & 8.4 & 67.6 & 63.5 & 52.0 & 36.1
\bb{50.7} & 21.5 & 22.0 & \bb{8.6} & \bb{69.7} & \bb{63.3} & \bb{52.0} & \bb{37.0}
% & 41.9 & 10.1 & 13.4 & 3.6\\
% & 42.9 & \bb{10.8} & \bb{14.3} & 4.0\\
& 43.7 & \bb{11.3} & \bb{14.6} & 4.3\\
\texttt{TsCA} (w filter)& \textbf{50.8} & \bb{21.7} & \textbf{22.3} & \textbf{8.7} & \textbf{69.8} & \textbf{63.4} & \textbf{52.2} & \textbf{37.1} 
% & 42.2 & 10.5 & 13.4 & 3.7 \\
& \bb{44.3} & \textbf{11.4} & \textbf{14.7} & \textbf{4.5} \\
% \texttt{TsCA}& \textbf{50.7} & 18.8 & 20.5 & 7.6 & 67.6 & \textbf{63.5} & \textbf{52.0} & \textbf{36.1} & 41.9 & 10.1 & 13.4 & 3.6\\
% \textbf{}

\bottomrule
\end{tabular}
\caption{
% Open-world results on MIT-States, UT-Zappos and C-GQA. \( S \) and \( U \) are the predict accuracies evaluated on seen and unseen compositions. \( H \) is the harmonic mean of \( U \) and \( S \) and AUC is the area under the curve.
CZSL comparisons in the open-world setting. We use `w' and `w/o' to distinguish models adopting filtering strategy to filter unfeasible compositions.
The best results are in \textbf{bold}. The second best results are in \bb{blue}.}
\label{tab:results_ow}
\end{table*}
%\vspace{-3mm}

% \begin{table}[t]
% \centering
%   \captionof{table}{Ablation results for each component on UT-Zappos in the closed-world setting.} % 使用captionof来添加标题
%   \label{tab:ablation}
%   \fontsize{6}{4.5}\selectfont % 设置字体大小和行距
%   \resizebox{\columnwidth}{!}{% 调整表格宽度为一栏的宽度
%     \begin{tabular}{ccccccccl}
%       \toprule
%        &$\mathcal{L}_{de}$& $\text{CT}_{\Cmat}$ & $\mathcal{L}_{cyc}$ &  S & U & H & AUC \\
%       \midrule
%       1 & &  &  & 65.7 & 73.5 & 54.4 & 40.8 \\
%       2 &\checkmark &   &  & 68.2 & 73.9 & 56.8 & 44.3 \\
%       % 2 &\checkmark &   &  & 67.7 & 73.7 & 56.6 & 43.2 \\
%       3 & & \checkmark  &   & 67.0 & 74.1 & 56.7 & 43.8 \\
%       4 & & \checkmark & \checkmark & 67.7 & 74.4 & 57.3 & 44.6 \\
%       % 4 &\checkmark & \checkmark  &   & - & - & - & - \\
%       5 & \checkmark & \checkmark  &  & 68.5 & 74.5 & 57.6 & 44.9 \\
%       \midrule
%       6 & \checkmark & \checkmark  & \checkmark  & 68.6 & 74.3 & 58.4 & 45.1 \\
%       \bottomrule
%     \end{tabular}
%   }
% \end{table}

\noindent \textbf{Metrics.} Following the common practice of prior works~\cite{lu2023decomposed}, we utilize the standard evaluation protocols and assessed all results using four metrics in both closed-world and open-world scenarios. Concretely, \textbf{S} measures the best seen accuracy when calibration bias is \(+\infty\) and \textbf{U} denotes the accuracy specifically for unseen compositions when the bias is \(-\infty\). To provide an overall performance measure on both seen and unseen pairs, we also report the area under the seen-unseen accuracy curve (\textbf{AUC}) by varying the calibration bias from \(+\infty\) to \(-\infty\) and identify the point that achieves the best harmonic mean (\textbf{H}) between the seen and unseen accuracy. Among these, \textbf{H} and \textbf{AUC} are the core metrics for comprehensively evaluating the model.  

\noindent \textbf{Implementation Details.} The proposed \texttt{TsCA} and all baselines are implemented with a per-trained CLIP ViT-L/14 model in PyTorch~\cite{paszke2019pytorch}. For the UT-Zappos, the hyper-parameters $\lambda_0$, $\lambda_1$, $\lambda_2$, $\lambda_3$ in losses are set as 1, 0.1, 10, and 0.1. For the MIT-States, the hyper-parameters are set as 1, 0.01, 0.1, and 0.01. For the CGQA, the hyper-parameters are set as 1, 0.01, 0.3, and 0.01. 
% During inference, the $\gamma$ is 0.8 for UT-Zappos and uniformly set to 0.4 for the rest datasets in the closed-world scenario, and is tailored to 0.4, 0.3, and 0.2, respectively, for each of the datasets in the open-world setting as mentioned above.
During inference, $\gamma$ is set to 0.8 for UT-Zappos and 0.4 for MIT-States and C-GQA in the closed-world scenario and is adjusted to 0.4, 0.3, and 0.2 for each of the datasets in the open-world setting. 
% The filtering threshold $T$ is set for all datasets.
% Following\cite{huang2024troika},we used a lightweight parameter efficient Adapter into the image encoder. 
% The primitive decoupler consists of two individual single-layer MLPs.
The primitive adapters consist of two individual single-layer MLPs.
All experiments are performed on a single 
NVIDIA RTX A6000 GPU. Please refer to the Appendix for more details. 

\subsection{Comparision with State-of-the-Arts}
We compare our \texttt{TsCA} with the most recent CZSL methods. The results are shown in Tab.~\ref{tab:results_cw} and Tab.~\ref{tab:results_ow}. On the closed-world setting, \texttt{TsCA} exceeds the previous SOTA methods on all datasets in core metrics. Specifically, it yields improvements of 0.6\% on MIT-States, 2\% on UT-Zappos, and 1.1\% on C-GQA in \textbf{H} over the second-best methods. It also attains considerable gains in \textbf{AUC} with increases of 0.5\%, 1.6\%, and 0.8\%, respectively. 
Notably, the advantage of UT-Zappos is more significant, suggesting that the fine-grained nature of UT-Zappos can better align with \texttt{TsCA}’s ability to achieve precise alignment across local visual features and textual representations. 
% which may be attributed to the fine-grained nature of the dataset, effectively aligning with \texttt{TsCA}'s capability for fine-grained alignment across local visual features and textual representations.
%likely because the fine-grained nature of the dataset aligns well with \texttt{TsCA}’s ability to achieve precise alignment across local visual features and textual representations. 
%
Similarly, the results in the open-world setting are also promising. 
%also demonstrate that our method achieves competitive performance. 
Our \texttt{TsCA} attains the highest \textbf{AUC} 
% across
% UT-Zappos and MIT-States
% , with the scores of 8.6\% and 36.7\%. 
across
all datasets
, with the scores of 8.7\% , 37.1\% and 4.5\%. 
It should be noted that the CT-based filtering strategy contributes to the improvements in this challenging setting. All numerical results substantiate our motivation to empower the model to capture semantic consistency among the image-composition-primitive interactions.

% \begin{table}[t]
% \centering
%   \fontsize{6}{7}\selectfont % 设置字体大小和行距
%   \resizebox{\columnwidth}{!}{% Adjust table width to column width
%     \begin{tabular}{l | c c c c}
%       \toprule
%       \rowcolor{gray!20} % 设置第一行的背景颜色为灰色
%       Components & S & U & H & AUC \\
%       \midrule
%       $\mathcal{L}_{base}$ & 65.7 & 72.5 & 54.4 & 40.8  \\
%       \midrule
%       + $\mathcal{L}_{de}$ &  66.5 & 73.9 & 56.2 & 43.5  \\
%       + $\mathcal{CT}$ & 67.0 & 74.1 & 56.7 & 43.8  \\
%       + $\mathcal{CT}$ + $\mathcal{L}_{cyc}$ & 67.7 & 74.4 & 57.3 & 44.6 \\
%       + $\mathcal{L}_{de}$ + $\mathcal{CT}$ & 68.5 & 74.5 & 57.6 & 44.9 \\
%       + $\mathcal{L}_{de}$ + $\mathcal{CT}$ + $\mathcal{L}_{cyc}$ & \textbf{68.6} & \textbf{74.3} & \textbf{58.4} & \textbf{45.1} \\
%       \bottomrule
%     \end{tabular}
%   }
%     \caption{Ablation results for each component on UT-Zappos in the closed-world setting.}
%     \label{tab:ablation}
% \end{table}
\begin{table}[t]
\centering
  \fontsize{7.5}{8.5}\selectfont
    \begin{tabular}{lp{0.25cm}p{0.25cm}p{0.25cm}p{0.3cm}p{0.25cm}p{0.25cm}p{0.25cm}p{0.3cm}}
      \toprule
      \multirow{2}{*}{Components} & \multicolumn{4}{c}{closed-world} & \multicolumn{4}{c}{open-world} \\
      \cmidrule(lr){2-5} \cmidrule(lr){6-9}
      & \cellcolor{gray!20}S & \cellcolor{gray!20}U & \cellcolor{gray!20}H & \cellcolor{gray!20}AUC & \cellcolor{gray!20}S & \cellcolor{gray!20}U & \cellcolor{gray!20}H & \cellcolor{gray!20}AUC \\
      \midrule
      $\mathcal{L}_{base}$ & 65.7 & 73.5 & 54.4 & 40.8 & 64.3 & 62.5 & 48.0 & 33.3 \\
      \midrule
      + $\mathcal{L}_{de}$ & 68.2 & 73.9 & 56.8 & 44.3 & 66.9 & 62.2 & 49.2 & 34.6 \\
      + $\mathcal{CT}$ & 67.8 & 74.3 & 57.5 & 44.6 & 67.0 & 64.6 & 50.1 & 35.9 \\
      + $\mathcal{CT}$ + $\mathcal{L}_{cyc}$ & 68.9 & 74.5 & 58.0 & 45.0 & 67.6 & 63.6 & 51.9 & 36.2 \\
      + $\mathcal{L}_{de}$ + $\mathcal{CT}$ & 69.0 & 75.4 & 58.2 & 45.3 & 67.9 & 63.7 & 52.0 & 36.3 \\
      + $\mathcal{L}_{de}$ + $\mathcal{CT}$ + $\mathcal{L}_{cyc}$ & 68.7 & 75.8 & 58.5 & 46.1 & 69.8 & 63.4 & 52.2 & 37.1 \\
      \bottomrule
    \end{tabular}
    \caption{Ablation results for each component on UT-Zappos.}
    \label{tab:ablation}
\end{table}
\vspace{-2mm}

\begin{figure*}
\centering
\includegraphics[width=0.9\textwidth]{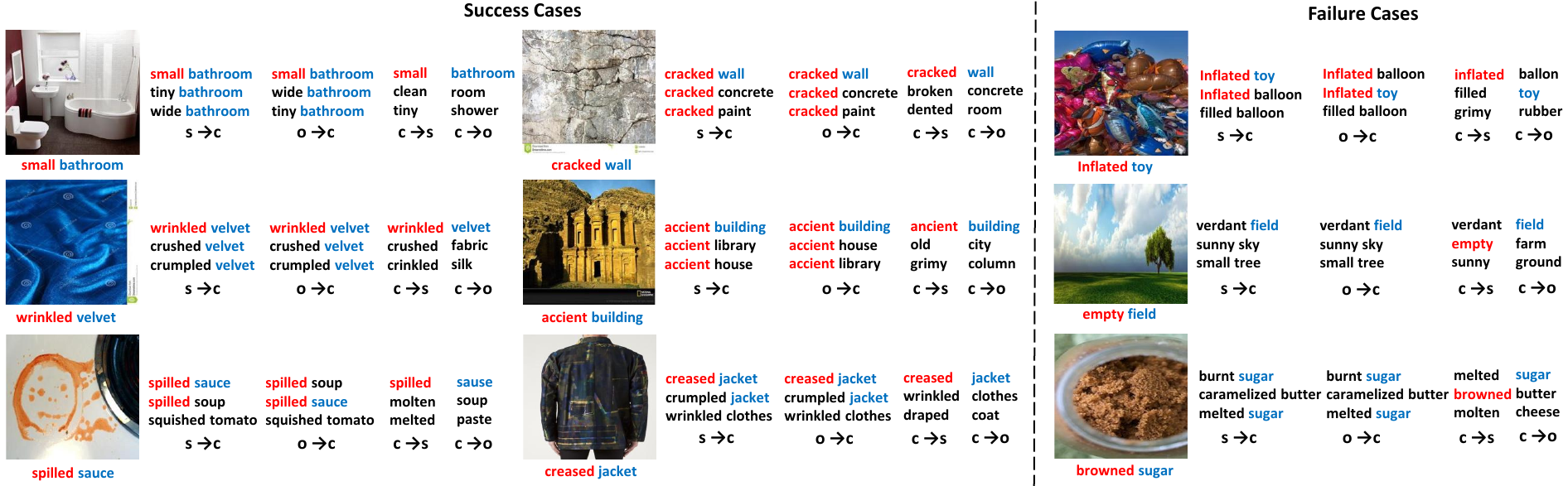}
\caption{\small{Qualitative results of the intra-modal transport plans on the MIT-States. For each sample, we show an image with the ground-truth composition, with the state indicated in red and the object in blue.  The top-3 predictions are presented in two formats: from the primitive class to the composition set in the first two columns, and from the composition label to the primitive set in the third and fourth columns. The annotations `\textbf{p}',`\textbf{s}', `\textbf{o}', and `\textbf{c}' correspond to patch, state, object, and composition, respectively.}}
\label{vis_intra}
\end{figure*}
%\vspace{-5mm}

\subsection{Ablation Study}
We empirically verify the effectiveness of each component in \texttt{TsCA} on UT-Zappos by comparing it against five variants. 
The results, seen in Tab. \ref{tab:ablation}, illustrate several observations: 1) The baseline model, i.e., removes the consistency-aware CT module and decoupler loss, shows the lowest performance. 2) Introducing either pairwise CT loss or decoupler loss on top of the baseline model significantly boosts all metrics. Notably, both primitive decoupler and cycle-consistency constraint positively contribute to pairwise CT. The former improves the quality of the textual sets, while the latter ensures semantic consistency during the transport chain, each leading to further enhancements in the model's effectiveness. 3) Combining the strengths of all components, our complete model achieves the best performance in terms of \textbf{H} and \textbf{AUC}.

\begin{figure}
\centering
\includegraphics[width=0.47\textwidth]{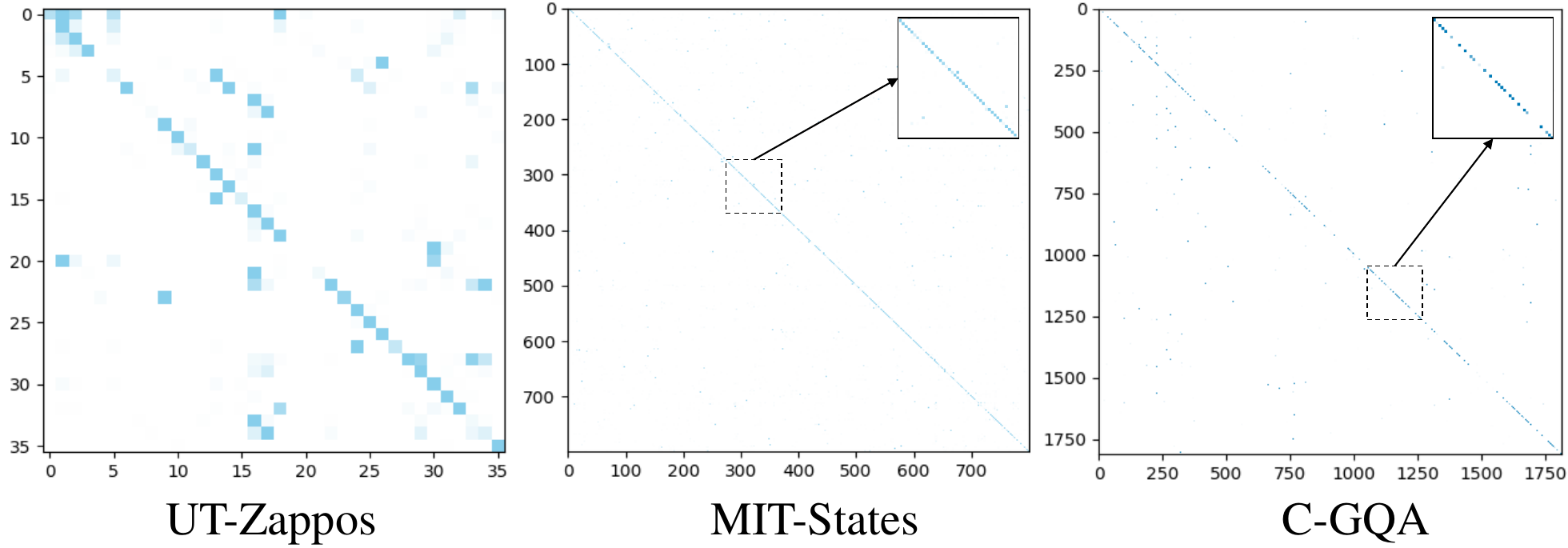}
\caption{\small{Cycle-consistency~(zoom-in for more details).}}
\label{cyc_diag}
\vspace{-2mm}
\end{figure}

To demonstrate the effectiveness of cycle-consistency, we draw the self-mapping matrix of the composition set for our full model on test data in Fig.~\ref{cyc_diag}. Though degradation occurs in the unseen pair positions, the results still exhibit a close similarity to the identity matrix. 
%reflecting that cycle-consistency is almost satisfied. 

\subsection{Qualitative Results}
\begin{figure}[t]
\centering
\includegraphics[width=0.4\textwidth]
{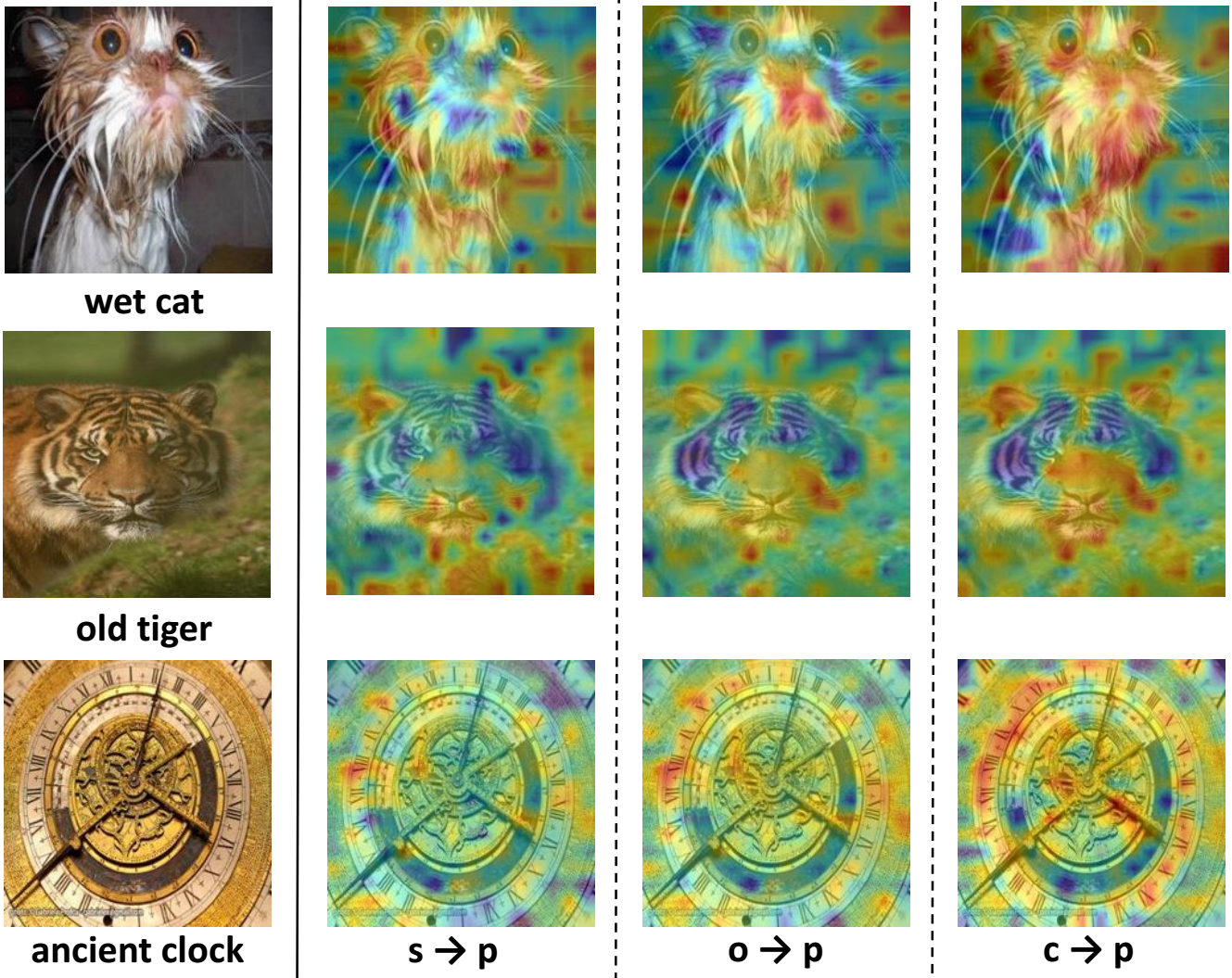}
\caption{\small{Visualization of the cross-modal transport plans on the Mit-States. Columns 1-3 represent the transport from the ground-truth state and object points in the primitive set, as well as the composition points in the composition set, to the patch set.}}
\label{vis_cross}
\vspace{-5mm}
\end{figure}

We further provide some visualization examples of transport plans learned in our \texttt{TsCA}. Recalling that the columns $\overleftarrow{\Tmat}_{21}$ and $\overleftarrow{\Tmat}_{31}$ depict how likely the corresponding textual semantics are transported to each visual patch. We convert each transport plan into heatmaps and resize them to combine with the raw image at Fig.~\ref{vis_cross}. We observe that different prompts from multi-path tend to align different patch regions, each of which contributes to the final prediction. 
% For example, in the first row, the state branch focuses more on the wet furs, the object branch considers the nose and mouse are important parts of recognizing a cat, and the composition branch  focuses more comprehensively on both.
For example, in the first row, the state branch emphasizes the wet fur, the object branch highlights the nose and mouth as key features for recognizing a cat, while the composition branch provides a more comprehensive focus on both aspects.

Moving beyond the visualization of cross-modal alignment, $\overleftarrow{\Tmat}_{32}$ and $\overrightarrow{\Tmat}_{23}$ also grant us access to intra-modal components, prompting an exploration of the interplay between compositions and primitives in an interpretable form. Fig.~\ref{vis_intra} presents the top-3 transport retrieval results for bidirectional transfers between the composition and primitive sets. Note that our model can both retrieve the correct composition from the primitive set, and also adeptly performs the inverse mapping of conceptual pairs. Besides, all top-3 retrieval results not only ensure the rationality of the state-object combination but also conform to the description of the images. 
% We also show some failures cases. Interestingly, despite being incorrect, we find that the retrieval labels can interpret the content of given images, thereby demonstrating the effectiveness of our \texttt{TsCA}. 
We also present some failure cases. Interestingly, even when incorrect, the retrieved labels still capture the content of the given images, demonstrating the effectiveness of our \texttt{TsCA}.

% This finding also meets with
% the motivation of the multi-mode prompt tuning, where each prompt aims to learn specific visual
% semantics.

\vspace{-2mm}

\section{Conclusion}
% In this paper, we rethink compositional generalization with a conditional transport perspective, upon the context of CZSL. 
In this paper, we revisit compositional generalization in CZSL through a conditional transport perspective. 
% Based on the multi-path paradigm, we first explored the pairwise CTs between the local visual feature set of images and two textual label sets. 
We explore pairwise CTs between the local visual features of images and two textual label sets within a multi-path paradigm. We then introduce cycle-consistency as a link to bond all sets, promoting robust learning. The primitive decoupler further improved accuracy and decoupled global primitive visual representations during prediction. Benefited from transport plan between primitive set and composition set, this approach successfully narrowed the composition search space by excluding unfeasible pairs during inference.
% Extensive experiments
% on three benchmarks consistently demonstrate the superiority
% of \texttt{TsCA}. Extensive ablation studies and visualizations confirm our motivation and the core role of each component. Since its natural flexibility and simplicity, we hope our work provides innovative ideas for follow-up researches.
Extensive experiments across three benchmarks consistently demonstrate the superiority of \texttt{TsCA}. 
Comprehensive ablation studies and visualizations confirm our motivation and the essential role of each component. 
With its inherent flexibility and simplicity, we hope our work inspires innovative ideas for future research.

\bibliographystyle{named}
\bibliography{ijcai25}

\end{document}